# THE EFFECT OF STRUCTURAL DIVERSITY OF AN ENSEMBLE OF CLASSIFIERS ON CLASSIFICATION ACCURACY

Lesedi Masisi, Fulufhelo V. Nelwamondo and Tshilidzi Marwala
School of Electrical and Information Engineering
University of the Witwatersrand
Private Bag 3
Wits
2050
South Africa

**ABSTRACT**

This paper aims to showcase the measure of structural diversity of an ensemble of 9 classifiers and then map a relationship between this structural diversity and accuracy. The structural diversity was induced by having different architectures or structures of the classifiers The Genetical Algorithms (GA) were used to derive the relationship between diversity and the classification accuracy by evolving the classifiers and then picking 9 classifiers out on an ensemble of 60 classifiers. It was found that as the ensemble became diverse the accuracy improved. However at a certain diversity measure the accuracy began to drop. The Kohavi-Wolpert variance method is used to measure the diversity of the ensemble. A method of voting is used to aggregate the results from each classifier. The lowest error was observed at a diversity measure of 0.16 with a mean square error of 0.274, when taking 0.2024 as maximum diversity measured. The parameters that were varied were: the number of hidden nodes, learning rate and the activation function.

## 1. Introduction

Developing an efficient way for classification has been a popular topic. It has been found that as opposed to using one classifier an ensemble of classifiers is more efficient [1]-[3]. The reason is that a committee of classifiers in making decision is better than one classifier. The individual classifiers which form this committee have created large interest when compared to accuracy of the ensemble [4]-[6]. Large research has been done in optimizing the diversity of the ensemble and the aggregation methods for the decision made by the ensemble [7]. This has led to developments in diversity measures and a relationship between these measures with the ensemble accuracy. Current methods use the outcomes of the individual classifiers of the ensemble to measure diversity [8]-[13]. These methods are applicable due to the way diversity was defined [14].

This study focuses on structural diversity. This means that, the individual parameters of the classifiers are used to measure structural diversity as opposed to viewing the outcome of the individual classifiers. This is in agreement with Sharkey [15], who stated that diversity can be induced by varying the architecture of the classifiers. It also further implies that diversity will not be induced by using different learning schemes such as bagging and boosting in sampling the data for training, this is done so that only the architectural parameters of the classifiers would induce diversity. Same data will be used to train the ensemble of classifiers. This will lead to knowledge on whether structural diversity has the potential to pose improvements on the classification.

There are a number of aggregation schemes such as minimum, maximum, product, average, simple majority, weighted majority, Naïve Bayes and decision templates to name a few, see [14], [16]. However for this study the majority vote scheme was used to aggregate the individual classifiers for a final solution. This report includes a section on the Identity structure (IDS), Kohavi-Wolpert Variance Method (KW), The neural network parameters, Genetical Algorithms (GA), The model, Implementation, Results and then lastly the conclusion and discussion.

## 2. Identity Structure (IDS)

The Identity Structure (IDS) is derived from taking into account the parameters that make up a Neural Network (NN). These parameters include the activation functions, number of hidden nodes and the learning rate. There are other types of the Neural Networks (NN) that can be used to form the IDS. A number of artificial machines can therefore be used for a hybrid ensemble. However this is beyond the scope of this study. For this study a Multi Layered Perceptron (MLP) was used. The parameters of concern were the number of hidden nodes, activation function and the learning rate. These parameters make up the Identity Structure of the classifiers (IDS). The IDS can be viewed as:

$$IDS = \begin{bmatrix} Machine\ type \\ Number\ of\ hidden\ nodes \\ Activation\ function \\ Learning\ rate \end{bmatrix}$$

The IDS was decrypted into a binary format, that contained 12 bits. That means a one would indicate that the parameter of the classifiers is active and a zero would mean the opposite. The first bit represented the machine type used, the five following bits represented the number of hidden nodes, the following three bits the activation functions and then lastly the last three bits represented the learning rates used by the classifier. Only three learnig rates were considered (0.01, 0.02, 0.03 and 0.04). That means a binary string of "0 0 1" would represent a 0.01 learning rate. Three activation functions were considered hence the three bits. These activation functions include the: Linear, Logistic and the Softmax. The first, second and third bit of the three bits represented the linear, Logistic and the Softmax respectively.

A one for the first bit of the decrypted IDS represented the MLP as the machine type used for that classifier. The number of hidden nodes is set not to exceed 30, hence five bits, see the five bold bits on the decrypted IDS below. This conversion makes the IDS less complex and would reduce the computational cost on the calculations for diversity. Suppose that the classifier was an MLP and had 5 hidden nodes and used a linear activation function and a learning rate of 0.02, then the IDS would be:

IDS = [1 **0 0 1 0 1** 1 0 0 0 1 0 ]

Each of the parameters of the IDS will have to be evaluated for measuring differences between the identities of the classifiers. The methods used to measure diversity are as follows: the Yule's Q-static for two classifiers, correlation coefficient ($\rho$), Kohavi-Wolpert variance (kw), Entropy measure (Ent), measure of difficulty ($\theta$) and Coincident Failure Diversity (CFD) [17], to name a few. These methods are mainly applied at the outcome of the classifiers and not at the building blocks (structure) of the classifiers [7]. However the Kohavi-Wolpert variance (*kw*) method can be applied to measure the structural diversity, which was derived from the variance formulation [17]. This is because diversity in this study is defined as the variance among the architectures of the individual classifiers.

## 3. Kohavi-Wolpert Variance Method (KW)

This method is applied in measuring the variance of the outputs of the classifiers in the ensemble. It falls under the family of Non-pairwise measures [7]. As mentioned above this equation is used to evaluate the outcomes of the classifiers. However, for this study it will be used to measure the variance of the different identities of the classifiers by evaluating the differences of the individual IDS of the classifiers. That means for this study:

$$l(V_j) = \sum_{i=1}^{L} D_{i,j} \qquad (1)$$

$V_j$, is a vector of the classifiers, L is the total number of classifiers. $V_j$ can be viewed as, $V_j = [ID_{1,j}^T, \ldots, ID_{L,j}^T]$. Equation (2) defines the overall variance calculation of the ensemble.

$$kw_r = \frac{1}{NL^2} \sum_{j=1}^{N} l(V_j)\big(L - l(V_j)\big) \qquad (2)$$

j = 1,…,N, where N is the number of the identity parameters (classifier type, complexity, activation function and the learning rate). This will result in the variance of the ensemble.

## 4. The Neural Network Parameters

The structural diversity is based on the parameters of the neural network. See Figure 1 for a MLP neural network.

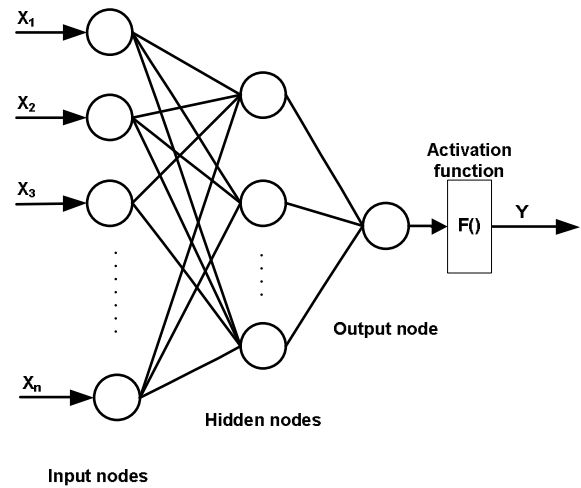

Figure 1: The MLP structure showing the inputs, the layers and the activation function

The MLP is composed of the input layer, hidden layer and the output layer, hence it is multi layered, see Figure 1. An MLP is built with different parameters, such as the activation functions, hidden nodes, biases, weights, etc. For this study diverse MLPs were created in a sense that they had different learning rates, activation functions and the number of hidden nodes. This would be considered as a diverse ensemble as compared to having the same MLPs with the same number of hidden nodes, activation function and the learning rate. This can clearly be seen from the IDS defined above.

It is clear that the diversity is not induced on the training of the neural network which is quiet a popular practice.

But diversity is derived from some of the building blocks of the individual classifiers. See equation (3) that describes the output of the neural network [18].

$$y_k = f_{outer}\left(\sum_{j=1}^{M} w_{kj}^{(2)} f_{inner}\left(\sum_{i=1}^{d} w_{ji}^{(1)} x_i + w_{jo}^{(1)}\right) + w_{ko}^{(2)}\right) \quad (3)$$

Where $f_{outer}$ and $f_{inner}$ are the activation functions at the output layer and at the hidden layers respectively, M is the number of the hidden units, d is the number of input units, $w_{ji}^{(1)}$ and $w_{kj}^{(2)}$ are the weights in the first and second layer respectively moving from input i to hidden unit j, and $w_{jo}^{(1)}$ indicate the biases for the unit j.

It was the outer activation functions which were varied to induce diversity. It can also be observed from equation (3) that varying the number of the hidden nodes will affect the generalization ability of the neural network.

## 5. Diversity and Genetical Algorithms (GA)

GA are evolutionary algorithms that aim to find a global solution to a given problem by applying the principles of evolutionary biology, such as mutation, crossover, reproduction and natural selection [20]. The GA have high capabilities to search large spaces for an optimal solution. The search process of the GA includes:

1. Generation of a population of offspring, normally taken as chromosomes

2. An evaluation function, that evaluates the fittest chromosome, if not fit genetic operations take over, such as: mutation and crossover. The mutation induces diversity in the search space.

3. This process continues until the fittest chromosome is attained.

However in this study the evaluation function is the diversity measure, the GA tries to meet a certain diversity (KW) among the ensemble of 9 classifiers, see Figure 2. The chromosomes are the indexes for the vector that contains 60 classifiers. The GA will then evolve the classifiers for a specified diversity value.

The GA faced difficulties in attaining the specified diversity. This was because the diversity measure specified could not be attained from the current ensemble of 60 classifiers. To prevent this problem from occurring one would need to:

- Build the ensemble of 60 classifiers with known KW values for any possible combination of the 9 ensembles.

- Initially run the GA for any KW values and then use the set of KW values that the GA can approximate. As the target values in the next run.

The second option seems much feasible than the first option because on the first option it would mean that there would be no need for the GA. The first option further implies that the GA would be synchronized with the KW measure. The GA was empirically optimized for an initial population of 20 chromosomes, 28 Generations with a crossover rate of 0.08.

## 6. The Model

The model describes the basic flow of the algorithm for developing an ensemble of 9 classifiers from the 90 classifiers. The method of voting was then applied on the 9 chosen classifiers for generating the classification accuracy of the ensemble.

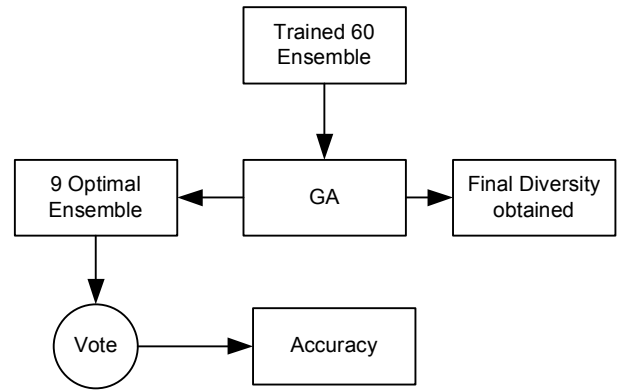

Figure 2: The mapping process of diversity and accuracy

## 7. Implementation

A vector of classifiers was created which was composed of 60 classifiers. This was because the more the classifiers there was, the better the search space for the GA for an optimal solution. All the classifiers in the vector were trained. The GA only looked for a solution for an ensemble of 9 classifiers. This means only 15% of the classifiers out of the ensemble of 60 classifiers was used at a time. This was to ensurereasonable diversity values in the search space. An odd number for the ensemble was chosen solemnly to avoid a tie when the method of voting was used.

The evaluation function was composed of two variables, the diversity measure and the targeted diversity ($T_{kwr}$). See equation (5) for the evaluation function.

$$f_{GA} = -(kw_r - T_{kwr})^2 \qquad (5)$$

Where: $f_{GA}$ is the evaluation function, $kw_r$ is the diversity measured of the 9 classifiers and $T_{kwr}$ is the targeted diversity.

The targeted diversity is the diversity the GA is searching for in the ensemble of 60 classifiers. That means the GA was searching for a group of 9 classifiers that would meet the targeted diversity. The structural diversity of the 60 ensemble was first calculated and was found to be 0.2024. Hence the targeted diversity value was ranged below this value (0.2024).

The GA tries to optimize the evaluation function by finding its maximum. Equation (5) will reach its maximum when the measured diversity is equal to the targeted diversity. GA was then optimized by first searching the KW values which the GA could nearly reach and then they were used in the second run as the target diversity values. This was to avoid the GA searching for the target values that did not exist from any combination of 9 classifiers from the 60 ensemble of classifiers.

### 7.1 Vector of classifiers

The classifiers were created via the normal distribution by creating them at random, the activation functions, hidden nodes, and the learning rate were chosen at random. This was so that the vector contained an ensemble of classifiers which were not biased. However a precaution was taken so that weak classifiers were not created, all the classifiers had the number of hidden nodes larger than the number of input features.

The vector also had classifiers that had a classification mean square error of less than 0.45 on the validation data set. For it was difficult to attain low mean square errors with the data used. The ensemble of 90 vectors was optimized by using an ensemble that produced a greater diversity measure. This diversity measure is 0.2024. This would be able to provide the GA with better classifiers that could generate the required diversity (KW).

### 7.2 The Nine Ensemble of Classifiers

The validation data set was used to select the nine classifiers from the vector of 60 classifiers. The classifiers were decrypted into a set of binary numbers as stated before. This binary number represented the IDS of the individual classifier. See table 1 for one of the ensemble of 9 classifiers.

Table 1: The IDS of the 9 classifiers

| C1 | C2 | C3 | C4 | C5 | C6 | C7 | C8 | C9 |
|----|----|----|----|----|----|----|----|----|
| 1  | 1  | 1  | 1  | 1  | 1  | 1  | 1  | 1  |
| 0  | 0  | 0  | 1  | 0  | 0  | 0  | 0  | 0  |
| 1  | 0  | 1  | 0  | 1  | 1  | 0  | 1  | 1  |
| 1  | 0  | 0  | 0  | 1  | 0  | 1  | 0  | 0  |
| 1  | 0  | 0  | 1  | 1  | 1  | 0  | 0  | 0  |
| 0  | 1  | 1  | 0  | 1  | 1  | 1  | 0  | 0  |
| 1  | 0  | 0  | 1  | 0  | 0  | 1  | 0  | 0  |
| 0  | 1  | 0  | 0  | 1  | 0  | 0  | 1  | 0  |
| 0  | 0  | 1  | 0  | 0  | 1  | 0  | 0  | 1  |
| 0  | 0  | 0  | 0  | 0  | 0  | 0  | 0  | 0  |
| 0  | 1  | 0  | 1  | 0  | 0  | 1  | 1  | 0  |
| 1  | 1  | 1  | 1  | 1  | 1  | 0  | 0  | 1  |

The maximum diversity given buy the ensemble of 60 classifiers was 0.2024, hence also the GA could not find any KW value beyond this point. This further limited the number of points that could be used to map the relationship between structural diversity and accuracy.

### 8. The Data

The interstate conflict data was used for this study. There are 7 features and one output, see table 1 for the data input features.

Table 1: The interstate conflict data

| inputs | values |
|--------|--------|
| Allies | 0-1 |
| Contingency | 0-1 |
| Distance | Log10(Km) |
| Major Power | 1-0 |
| Capability | Log10 |
| Democracy | -10-10 |
| Dependency | continuous |

The output is a binary number, a zero represented no conflict where else a one represented conflict. There are a total of 27,737 cases in the cold war population. The 26,846 are the peaceful dyads year and 875 conflict dyads year [19]. This shows clearly that the data is complicated for training a neural network. However for this study it was just used to show how structural diversity relates with the ensemble accuracy.

A data sample of 1006 was used for training, 317 samples for validation and 552 for testing. The total data used was therefore 1875. This data has seven feature inputs as mentioned, however the data was normalized between 0 and 1, to have equal importance of all the features, by using equation (4):

$$X_{norm} = \frac{x_i - x_{min}}{x_{max} - x_{min}} \qquad (4)$$

Where $x_{min}$ and $x_{max}$ are the minimum and maximum values of the features of the data samples observed, respectively.

## 9. Results

Figure 3 shows the results from the GA with the first run of the GA with arbitrary target values. However Figure 4 shows the graph of error Vs structural diversity with the optimized target values. The figures were obtained from using the validation data set. The ensemble of 9 classifiers chosen by the GA was then tested on the testing data set so as to bring more sense to the results, see table 2. The testing data was applied on the ensemble that produced 0.16 and 0.11 diversity values.

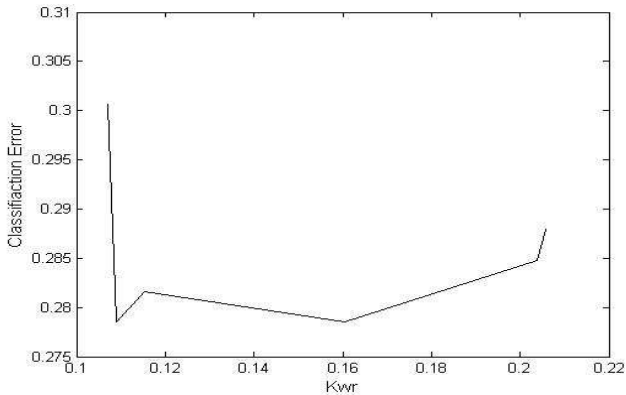

Figure 3: GA predicting 6 diversity values

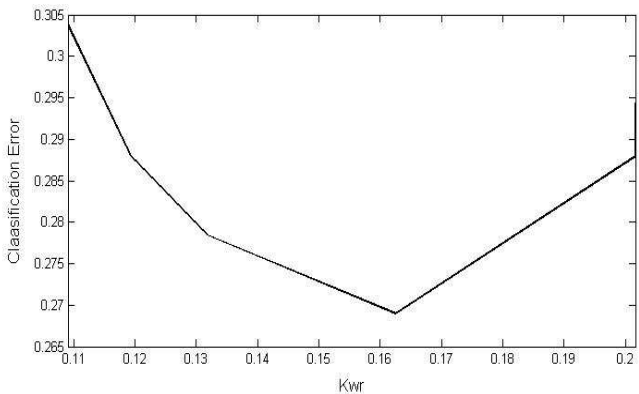

Figure 4: Optimised GA on the same 60 ensemble.

Table 2: Classification error on the testing data set

| Kw | Error (Initial Kw) | Errors (Optimized Kw) |
|---|---|---|
| 0.11 | 0.3128 | 0.2821 |
| 0.16 | 0.2821 | 0.2749 |

It can be seen that the results from Figure 3 and Figure 4 follow the expected trend. The error decreased with increasing diversity. However there is a point where the degree of diversity becomes unfavorable. The error began to increase with an increase in diversity. This is in alignment with [7], who stated that diversity can either profit the system or it could bring about poor performance on the classification. It can also be observed from the graphs that the data points of interest are not to scale. The occurrence of a change is not consistent. This is could be attributed to:

- The fact that there was a lot of rounding off values in the software package (Matlab),

- The other factor is that the ensemble of 60 classifiers was not designed with a linear or with consistent increments of diversity values.

- The targeted diversity values might not have been possible to be extracted from the ensemble and due to that the GA will provided its local solution.

Mean square error was used in all instances to calculate the classification accuracy. However it was just used as reference so as to observe the behavior of the ensemble with the changing diversity measured.

## 10. Conclusion and discussion

This paper presented a measure of structural diversity as defined in this paper and then a relationship between structural diversity and classification accuracy were mapped. As diversity increases the generalization ability of the ensemble improved, this was seen by the classification error decreasing. However there was a point where diversity made the ensemble weaker to classify. This study has also shown that diversity of an ensemble can be induced by having an ensemble that is composed of classifiers that have different parameters such as activation functions, number of hidden nodes and the learning rate. This is in alignment with Sharkey [15]. The methods used were computationally expensive since they made use of the GA and the training of 60 classifiers. This study agrees with most literatures that diversity does improve the accuracy of the ensemble [7]. This was observed by using the testing data set on the ensemble that had a low classification error. This study was limited by the bank of classifiers (60 classifiers) that were created at random. This ensemble had 0.2024 diversity measures which meant that only small samples could be used to verify the relationship between diversity and accuracy. All the errors on the testing data set showed that diversity can be used to measure the potential for improvement on the ensemble of classifiers.

## References

[1] J. Kittler, M. Hatef, R. Matas & J. Duin, On combining classifiers. *Intell.* 20 (3), 1998, 226-239.

[2] L. Breiman, Combining predictors in: A.J.C Sharkey (Ed.), *Combining Artificial Neural Nets*, Springer, London, 1999.

[3] H. Drucker, Boosting using using neural networks, in: A.J.C. Sharkey (Ed.), *Combining Artificial Neural Nets.* Springer, London, 1999.

[4] F. Giacint & G. Roli, Hybrid methods in pattern recognition, *World Scientific Pres,* Singapore, 2002.

[5] K. Sirlantzis, S. Hoque & M.C. Fairhurst, Diversity in multiple classifier ensembles based on binary feature quantisation with application to face recognition, *Department of Electronics, University of Kent*, United Kindom, 2008, 437-445.

[6] G. mordechai, J.H. May & W.E. Spangler, Assessing the predictive accuracy of diversity measures with domain-dependent, asymetric misclassification costs, *Information Fussion 6*, 2005, 37-48.

[7] L.I. Kuncheva, M. Duin & R.P.W. Skurichina, An experiment study on diversity for bagging and boosting with linear classifiers, *Information Fusion 3*, 2002, 248-250.

[8] K.M. Ali & M.J. Pazzani, Error reduction through learning multiple descriptions, *Machine Learning 24 (3)*, 1996, 173-202.

[9] P.K. Chan & D.J. Stolfo, On the accuracy of meta-learning for data minin,. *Jornal of Intelligent Information System 8,* 1997, 5-28.

[10] T.H. Davenport, Saving IT's soul: human-centered information management, *Harvard Business Review*, 1994, 119-131.

[11] T.G. Dietterich, An experimental comparison of three methods for constructing ensembles of decission trees: bagging and boosting, and randomization, *Machine Learning 40* (2), 2000, 139-157.

[12] l. Hansen & P. Salamon, Neural network ensemble, *IEEE Transaction on Partten Analysis and Machine Intelligence 12* (10), 1990, 993-1001.

[13] A. Krogh & J. Vedelsby, Neural network ensembles cross validation and active learning, in: G. Tesauro, D.S. Touretzky, T.K. Leen (Eds.), *MIT Press,* Cambridge, 1995.

[14] A.C. Shipp & L.I. Kuncheva, Relationship between combination methods and measures of diversity in combining classifiers, *Elsevier Science B.V.*, UK, 2002, 135-148.

[15] A. Sharkey, Multi-Net systems, Combining artificial neural nets: Ensemble and Modular Multi-net Systems, *Springer-Verlag*, 1999, 1-30.

[16] K.Tumer & J. Ghosh, Linear and order statistics combiners for pattern classification, *Combining Artificial Neural Nets*, London, 1999.

[17] R. Kohavi & D.H. Wolpert, Bias plus variance decomposition for zero-one loss functions, Morgan Kaufmann, 1996, 275-283.

[18] C.M. Bishop, *Neural networks for pattern recognition (* Oxford University Press, 1995).

[19] T. Marwala & M. Lagazio, Modeling and Controlling Interstate Conflict.

[20] S. Mohamed, Dynamic Protein Classification: Adaptive Models Based on Incremental Learning Strategies. *University of the Witwatersrand*, MSc theses. 2006.